\pdfoutput=1

\documentclass[11pt]{article}

\usepackage{acl}

\usepackage{times}
\usepackage{latexsym}

\usepackage[T1]{fontenc}

\usepackage[utf8]{inputenc}

\usepackage{microtype}

\usepackage{inconsolata}

\usepackage{comment}
\usepackage{booktabs}
\usepackage{graphicx}
\usepackage{multirow}
\graphicspath{{figs/}}
\usepackage{lipsum}

\usepackage{bbm}
\usepackage{amsmath}
\usepackage{float}
\usepackage{subcaption}
\usepackage{booktabs}
\usepackage{cleveref}

\usepackage{makecell}
\usepackage{mathtools}
\usepackage{amssymb}
\usepackage{siunitx}
\usepackage{etoolbox}           
\renewcommand{\bfseries}{\fontseries{b}\selectfont} 
\robustify\bfseries             
\newrobustcmd{\B}{\bfseries}    
\title{Fine-Grained Prediction of Reading Comprehension from Eye Movements}

\author{Omer Shubi$^1$, Yoav Meiri$^1$, Cfir Avraham Hadar$^1$, Yevgeni Berzak$^{1,2}$ \\
 $^1$Faculty of Data and Decision Sciences, \\
 Technion - Israel Institute of Technology, Haifa, Israel \\
 $^2$Department of Brain and Cognitive Sciences, \\
 Massachusetts Institute of Technology, Cambridge, USA \\
 \texttt{\{shubi,meiri.yoav,kfir-hadar\}@campus.technion.ac.il},
\texttt{berzak@technion.ac.il} \\
}

\begin{document}
\maketitle
\begin{abstract}

Can human reading comprehension be assessed from eye movements in reading? In this work, we address this longstanding question using large-scale eyetracking data. We focus on a cardinal and largely unaddressed variant of this question: predicting reading comprehension of a single participant for a \emph{single question} from their eye movements over a \emph{single paragraph}. We tackle this task using a battery of recent models from the literature, and three new multimodal language models. We evaluate the models in two different reading regimes: ordinary reading and information seeking, and examine their generalization to new textual items, new participants, and the combination of both.  The evaluations suggest that the task is \emph{highly challenging}, and highlight the importance of benchmarking against a strong text-only baseline. While in some cases eye movements provide improvements over such a baseline, they tend to be small. This could be due to limitations of current modelling approaches, limitations of the data, or because eye movement behavior does not sufficiently pertain to fine-grained aspects of reading comprehension processes. Our study provides an infrastructure for making further progress on this question.\footnote{Code is available at \href{https://github.com/lacclab/Reading-Comprehension-Prediction}{https://github.com/lacclab/Reading-Comprehension-Prediction}.}

\end{abstract}

\section{Introduction}
\label{sec:intro}
Reading comprehension is an indispensable skill for successful participation in modern society. Consequently, many efforts and resources are invested in the development of reading comprehension assessments by educational institutions and commercial companies. The standard, and to date the only practical way to assess reading comprehension is through behavioral tasks, most commonly reading comprehension questions. However, despite its clear value and ubiquitous use, this approach is extremely time-consuming and costly, which severely limits the volume and public availability of reading comprehension tests. Further, this testing methodology relies on \emph{offline} behavioral signals -- the end responses to a few select reading comprehension questions, and has no ability to trace the rich \emph{online} reading comprehension processes as they unfold over time.

An alternative vision for assessing reading comprehension has been emerging in psycholinguistics and the psychology of reading. It posits that reading comprehension may be decoded in real-time directly from eye movements in reading. This vision is rooted in literature that suggests a tight correspondence between eye movements and real time language comprehension \citep[among others]{JustCarpenter1980,Rayner1998,rayner2016}. With the rise of modern machine learning and NLP, multiple studies over the past decade attempted to use eye movement data to predict reading comprehension \cite[][among others]{copeland2014,Ahn2020SBSAT,ReichETRA2022,meziere2023using}. This line of work suggests that in some cases various aspects of reading comprehension can be predicted from eye movements with above-chance performance. However, despite the advances so far, predictive modeling of reading comprehension from gaze is still in its infancy. 

A number of factors have been hindering progress in this area. One is the paucity and small size of reading comprehension data paired with eye movements. 
Second, the task of reading comprehension prediction has thus far been predominantly formulated as prediction of \emph{aggregated scores across multiple questions} rather than prediction of comprehension at the resolution of an individual question. 
Further, reading comprehension has been primarily studied when the reader has no specific goals with respect to the text beyond general comprehension, a regime that we refer to as \emph{ordinary reading}. Many other reading regimes common in daily life, such as explicit information seeking, remain largely unaddressed. Finally, despite the dramatic progress in machine learning and NLP in recent years, effective joint modeling of text and eye movements  remains a nascent and challenging domain of investigation.

In this work, we take a step forward in advancing the state-of-the-art in eye movement-based prediction of reading comprehension by combining new models, new data, and systematic evaluations. 
Our primary contributions are the following:
\begin{itemize}

    \item \textbf{Task}: we introduce the challenging and largely unaddressed task of predicting the reading comprehension of a single reader with respect to a \emph{single reading comprehension question over one passage}. This task is enabled by OneStop Eye Movements \citep{malmaud_bridging_2020}, the largest eyetracking for reading comprehension dataset to date with 486 multiple-choice questions and 19,440 question responses from 360 participants.
    
    \item \textbf{Modeling}: we develop three new models that combine text and eye movements based on the transformer encoder architecture: RoBERTa-QEye, MAG-QEye, and PostFusion-QEye. These models address both test format-agnostic and multiple-choice specific variants of the task.
    
    \item \textbf{Reading Regimes}: we study reading comprehension not only in ordinary reading but also in information seeking, a highly common but understudied reading scenario.
    
    \item \textbf{Evaluation}: we evaluate our models against a battery of existing models for prediction of reading comprehension from eye movements, and a strong text-only baseline. To this end, we use a detailed evaluation protocol targeting three different levels of model generalization: new participant, new textual item, and the combination of both.  
    
\end{itemize}

\section{Related Work}
\label{sec:related-work}

Our study contributes to an existing body of work on the prediction of reading comprehension from eye movements in reading. 
To address various aspects of this task, prior studies used a wide range of models, including linear models \cite{meziere2023using, meziere2023scanpath}, kernel methods \cite{makowski2019}, feed-forward networks \cite[e.g.][]{copeland2014}, CNNs \cite{Ahn2020SBSAT} and RNNs \cite[e.g.][]{Ahn2020SBSAT,ReichETRA2022}. These were typically applied to the prediction of aggregated comprehension scores over multiple items. In this work, we evaluate multiple models from prior work on the single-item reading comprehension task. 

While transformer models \cite{vaswani_attention_2017}, have been used for joint modeling of eye movements and text \cite[e.g.][]{Deng_Eyettention2023,yang_plm-as_2023}, they have not been applied to the problem of reading comprehension prediction from eye movements. In this work we introduce three new transformer models which draw on  
multi-modal transformers, in particular MAG \cite{rahman_integrating_2020} which integrated text, speech and vision for sentiment analysis, and language vision models such as VisualBERT \cite{li_visualbert_2019} (see \citet{zhu_multimodal_2023,xu_multimodal_2023} for reviews).

Most prior studies on reading comprehension prediction from eye movements relied solely on eye movement features \cite{copeland2014,southwell2020,Ahn2020SBSAT,meziere2023using,meziere2023scanpath}, while a few combined eye movements with properties of the underlying text \cite{martinez2014,makowski2019,ReichETRA2022}. In the current work, we take the latter, under-explored approach. The importance of combining eye movements with attributes of the text is motivated by a large literature in the psychology of reading which points to systematic effects of linguistic properties of the text on reading times \cite[among  others]{Rayner1998,rayner2004effects,kliegl2004effects,dembergkeller2008,smithlevy2013}, in particular in the context of reading comprehension \cite{JustCarpenter1980} and linguistic proficiency \cite{berzak_assessing_2018,berzak2023traces}. 

While highly informative, existing work is critically limited by small data, especially with respect to the number of available questions and participants. For example, \citet{copeland2014} have 9 text pages, 18 questions and 39 participants. 
SB-SAT \cite{Ahn2020SBSAT}, the only publicly available eyetracking dataset for reading comprehension, has 22 text pages, 20 questions, and 95 participants. 
The small size of previously used datasets severely limits the potential of NLP and machine learning approaches for reading comprehension prediction. 
At the same time, the reading comprehension component of broad coverage eyetracking datasets such as MECO \cite{meco2022} and CELER \cite{celer2022} comprises only simple comprehension questions that serve as attention checks, and as such are not well suited for studying reading comprehension. OneStop, used here, has a large number of items, participants and questions, enabling meaningfully addressing item-level prediction of comprehension. 

Prior work varies in experimental designs. In several studies, multiple questions are presented after reading a multi-screen text without the ability to return to the text \cite{makowski2019,Ahn2020SBSAT,ReichETRA2022}. This design is advantageous in the separation of text reading and question answering, but can lead to loose relations between eye movements and question-answering behavior due to memory limitations. In other studies, such as \citet{copeland2014}, participants can switch back and forth between the text and the questions. This creates a complex mix of ordinary reading and information seeking components which are difficult to disentangle.  
In OneStop, a single question appears immediately after reading a single text page, setting a middle ground between the two primary existing approaches for question presentation, and alleviating their main disadvantages. At the same time, it includes a question preview manipulation which allows to systematically compare reading comprehension in ordinary reading and question guided information seeking.

An additional limitation of prior work is the scope and nature of the evaluations. With the exception of \citet{copeland2014}, both training and evaluation were previously carried out over \emph{aggregated responses} across multiple questions, and in some cases also across multiple texts. These approaches, which focus on measuring overall comprehension, do not enable testing direct links between eye movements and understanding specific aspects of the text. In several studies \cite{martinez2014,makowski2019,Ahn2020SBSAT,ReichETRA2022}, an additional step was taken, binning comprehension scores into two binary categories, high versus low comprehension, thus further simplifying the task. 

A second important evaluation limitation in prior work is evaluations in which eyetracking data for both the test participants and items is used in the training set. To our knowledge, except for \citet{makowski2019}, no work has evaluated reading comprehension prediction when neither the participant nor the item appears in the training data. This evaluation regime is needed to fully characterize model generalization ability. 
Importantly, even in less challenging regimes and with aggregated scores and binning, model performance in prior work is typically only modestly higher than chance level. More stringent evaluations without binning comprehension scores \cite{martinez2014}, or with held-out participants and/or items \cite{makowski2019,ReichETRA2022} tend to exhibit chance level performance. These results suggest that generalization in reading comprehension prediction is highly challenging. 

\section{Eyetracking Data}
\label{data-and-annotations}

We use OneStop, an extended version of the dataset collected by \citet{malmaud_bridging_2020} over the textual materials of OneStopQA \cite{berzak-2020-starc}. OneStop is the largest English L1 eyetracking for reading corpus to date. The data was collected using an Eyelink 1000+ eyetracker at a sampling rate of 1000Hz. In this dataset, 360 adult native English participants read newswire articles from the Guardian, and answer a multiple-choice reading comprehension question about each paragraph. The dataset includes 30 articles divided into 162 paragraphs. The average paragraph length is 109 words. Each paragraph has 3 possible questions, corresponding to a total of 486 questions. 

\begin{table}[ht]
\centering
\resizebox{\columnwidth}{!}{%
\begin{tabular}{@{}lllllcc@{}}
\toprule
 \textbf{Answer} & \textbf{Category} & \textbf{Degree of Comprehension} & \textbf{Gathering} & \textbf{Hunting} \\ \midrule
$A$ & Correct & Full comprehension & 7,890 (81.2) & 8,450 (86.9) \\
$B$ & Incorrect &\makecell[l]{Identified question-relevant information} & 1000 \ (10.3) & 744 
 \ \ \  (7.7) \\
$C$ & Incorrect & \makecell[l]{Some degree of attention to the text} & 568 \ \ \ (5.8) & 374 \ \ \  (3.8) \\
$D$ & Incorrect & No evidence for comprehension & 260 \ \ \ (2.7) & 152 \ \ \ (1.6) \\ \bottomrule
\end{tabular}%
}
\caption{Summary of the STARC annotation framework for answer types $A$--$D$, their corresponding degree of comprehension, and number of trials in which each answer type was chosen in OneStop. Values in parentheses are percentages by reading regime.}
\label{tab:starc}
\end{table}

\begin{figure*}[ht!]
    \centering
\includegraphics[width=\linewidth]{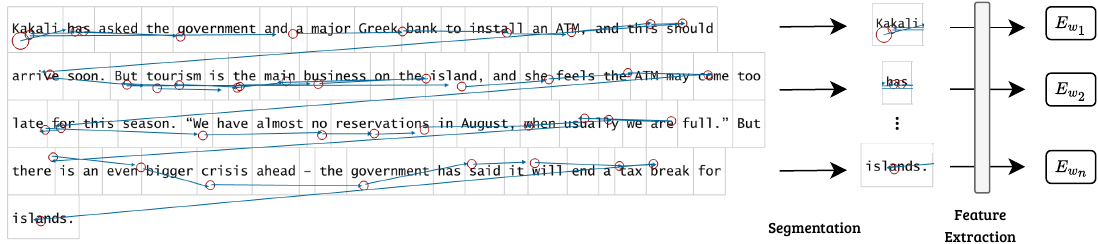}
    \caption{Left: an example of an eye movement trajectory over a paragraph, where red circles represent fixations, and blue arrows represent saccades. Right: a schematic depiction of word-level feature extraction, resulting in a vector $E_{w_i}$: an eye movements and linguistic word properties feature representation for each word.}
    \label{fig:input}
\end{figure*}

The articles are divided into three 10-article batches, where each participant is assigned to one batch. In each trial of the experiment, participants read a paragraph and then proceed to answer one of the three possible questions on a new screen, without the ability to return to the paragraph. 180 participants are in an ordinary reading (Gathering) regime where they do not see the question prior to reading the paragraph. The remaining 180 participants are in an information seeking regime (Hunting) where they are presented with the question (but not the answers) before reading the paragraph. The total number of trials is 19,440, split equally across the two reading regimes. This corresponds to 40 responses per question, 20 for each regime--paragraph combination. The total number of word tokens over which eyetracking data was collected in OneStop is 3,827,216.  

The underlying textual materials and reading comprehension questions follow the STARC annotation framework \cite{berzak-2020-starc}, 
where answer $A$ is the correct answer, answer $B$ is a miscomprehension of the information required to answer correctly, $C$ refers to another part of the text that does not provide the answer to the question and $D$ has no textual support. These answer types correspond to an ordering of the answers by degree of comprehension. Table~\ref{tab:starc} presents a summary of the framework along with answer choice statistics in the OneStop eyetracking data.

\section{Tasks}
\label{sec:problem-setting}

\subsection{Correct versus Incorrect Comprehension}

The primary task we address is item-level prediction of whether a participant will respond correctly to a single question about a paragraph from the participant's eye movements over the paragraph. For each paragraph $p$ and a corresponding question $q^p$, the possible answers are $Ans^{q^p}=\{a^{q^p}_1,a^{q^p}_2,a^{q^p}_3,a^{q^p}_4\}$. Note that the correct answer $A$ and the three distractors $\{B, C, D\}$ are randomly mapped per trial to $a_1$ through $a_4$. 
The set of $p$, $q^p$, and optionally $Ans^{q^p}$, defines a \textit{textual item} $W$. Given a participant $S$ tested on item $W$, where the participant's eye movements over the paragraph are $Eyes^p_S$, the complete trial information is $Trial^W_S \coloneq \{W, Eyes^S_p\}$. We make $W$ optional to allow for models that use only eye movements without the text. 

The prediction problem can then be formulated as a binary classification task, we predict whether the participant will answer the question correctly. Formally, given a classifier $h$:
\begin{equation} \label{eq:task-def}
    h: Trial^W_S \mapsto \{0,1\}
\end{equation}
where $1$ indicates a correct answer ($A$) and $0$ indicates an incorrect answer $(B/C/D)$.

Note that this task formulation abstracts away from the multiple-choice format. This allows assessing comprehension without depending on the format of the subsequent assessment task (e.g. answer choice, answer production), nor its details such as the number of answer choices and their specific content in the multiple-choice format. 
The combination of these task characteristics enables applying prior models from the literature, all of which predict a binary outcome without taking into account the answers, and some of which use only eye movements without the text. 

\subsection{Specific Answer Choice}
We further address a task that takes advantage of the multiple-choice assessment format. In this task, given the answers, we predict which \textit{specific answer} the participant will select:
\begin{equation} \label{eq:task-def2}
    h: Trial^W_S \mapsto \{a_1,a_2,a_3,a_4\}
\end{equation}


\section{Models}

\begin{figure*}[ht!]
    \centering
        \begin{subfigure}[b]{0.28\textwidth}
        \centering
        \includegraphics[height=6cm]{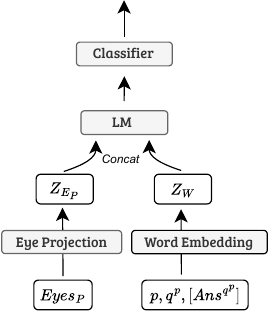}
        \caption{RoBERTa-QEye}
        \label{fig:bertqeye}
    \end{subfigure}
    \hspace{0.05\textwidth}
    \begin{subfigure}[b]{0.28\textwidth}
        \centering
        \includegraphics[height=5cm]{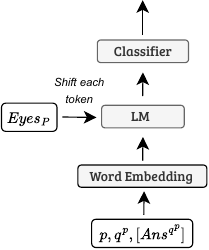}
        \caption{MAG-QEye}
        \label{fig:mag}
    \end{subfigure}
    \hspace{0.05\textwidth}
    \begin{subfigure}[b]{0.28\textwidth}
        \centering
        \includegraphics[height=6cm]{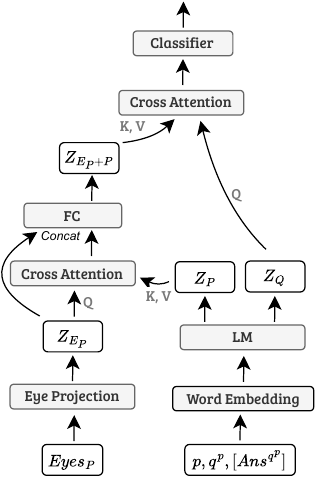}
        \caption{PostFusion-QEye}
        \label{fig:postfusion}
    \end{subfigure}
    \caption{Model architectures. (a) RoBERTa-QEye treats eye movements as additional input features. (b) MAG-QEye uses eye movement information to modify contextualized word representations. (c) PostFusion-QEye processes text and eye movements separately and combines them via cross-attention mechanisms. Model input: $Eyes^P$ represents the participant's eye movements over the paragraph $p$, $q^p$  is a question and $[Ans^{q^p}]$ are optional answer choices which are provided only in the multiple choice version of the task.}
    \label{fig:models}
\end{figure*}

We introduce three new models, RoBERTa-QEye, MAG-QEye and PostFusion-QEye, all of which combine text and eye movements information, and rely on the transformer language model encoder. Specifically, we use the $\text{RoBERTa}_{\text{LARGE}}$ model \cite{liu_roberta_2019}. Each of these models uses a different strategy for combining text with eye movements. RoBERTa-QEye augments the textual input with additional eye movement features. MAG-QEye uses eye movement information to modify contextualized word representations at intermediate layers of the language model. PostFusion-QEye processes text and eye movements separately and then combines them via cross-attention mechanisms. We further adjust a number of prior models from the literature for the single-item reading comprehension prediction task.

\textbf{Eye Movement Feature Representations} The eyetracking record is commonly represented as a scanpath consisting of fixations (periods in which the gaze position is stable) and saccades (rapid transitions between fixations). The examined models represent this information in three different ways, in increasing level of granularity:
\begin{itemize}
    \item \textbf{Global}: Summarizing fixation and saccade information across all the words in the input.
    \item \textbf{Words}: Summarizing fixation and saccade information for each word.
    \item \textbf{Fixations}: Accounting for each fixation and its preceding and following saccade.
\end{itemize}

Our new models focus on the word and fixation level approaches, using a variety of eye movement measures from the psycholinguistic literature.  As reading times are known to be affected by linguistic word properties such as predictability, frequency, and length 
\cite{rayner2004effects,kliegl2004effects,rayner2011}, which are not directly encoded in word embeddings, we further add such properties to the eye movement representations to allow the models to learn eye movements-word property interactions. The strength of such interactions has been shown to be indicative of the readers' linguistic proficiency \cite{berzak_assessing_2018,berzak2023traces}, which is directly related to reading comprehension. The eye movement and linguistic word property features used in all the models are listed in \Cref{sec:app-eye-features}. Note that two different feature sets are used for representing eye movements at the word and fixation levels. Figure \ref{fig:input} presents an example of an eye movement trajectory over a paragraph and a schematic visualization of the word-level feature extraction approach.

\subsection{RoBERTa-QEye}

RoBERTa-QEye incorporates eye movements as additional input sequences to RoBERTa by projecting them to the word embedding space. An overview of the architecture is presented in \Cref{fig:bertqeye}. The model is implemented in two variants, RoBERTa-QEye-Words which has a word-level feature representation and RoBERTa-QEye-Fixations, which uses a fixation-level representation. Both variants combine a textual input $Z_W$ with eye movements input $Z_{E_P}$.

The textual representation $Z_W$ is the word embedding sequence $[\texttt{CLS}; p; \texttt{SEP}; q^p; [Ans^{q^p}]; \texttt{SEP}]$, where $p$ is the paragraph, $q^p$ is the question, $[Ans^{q^p}]$ are optional answers, and $\texttt{SEP}$ is a separator token. The eye movement representation for the paragraph $Z_{E_P} = [Z_{E_{w_1}},...,Z_{E_{w_n}}]$ consists of a representation for each fixation or word $i$ as: \begin{equation}
    Z_{E_{w_i}} = \text{FC}(E_{w_i}) + \text{Emb}_{\text{pos}}(i) + \text{Emb}_{\text{eye}}
    \end{equation}
    where $E_{w_i}$ are the eye movement and word property features and FC is a fully connected layer projecting this feature representation to the word embedding space. $\text{Emb}_{\text{pos}}(i)$ is the positional embedding of the $i$-th word or fixation, initialized to the model's original positional embedding, which ties the eye movement representation to its respective word index. $\text{Emb}_{\text{eye}}$ is an additional learnable embedding marking the presence of eye movement information. $Z_{E_P}$ is concatenated with the word embedding representation $Z_W$, separated by a special token $\texttt{SEP}_E$, initialized as $\texttt{SEP}$. The combined sequence $[Z_{E_P}; \texttt{SEP}_E; Z_W]$ is passed through the transformer encoder language model. The resulting \texttt{CLS} token is then provided to a multilayer perceptron for response prediction. 

\subsection{MAG-QEye}

MAG-QEye, shown in \Cref{fig:mag}, modifies the transformer encoder's hidden word representations based on eye movement information. It is an adaptation of the MAG architecture \cite{rahman_integrating_2020} originally developed for multimodal sentiment analysis. The goal of this model is to emphasize or de-emphasize words based on their respective eye movement features.  Formally, for a given model layer $k$, each hidden token representation in the paragraph $Z^k_{W_i}$ is shifted by $H_{W_i}$:
\begin{equation}
    \bar{Z^k}_{W_i} = Z^k_{W_i} + \alpha H_{W_i}
\end{equation}
where $H_{W_i}$ is a scaled version of eye movements $E_{W_i}$ transformed into the word embedding space. 
The final resulting \texttt{CLS} token is passed through a multilayer perceptron classifier.
\Cref{sec:app-baselines-mods-mag} provides a detailed description of the architecture.

\subsection{PostFusion-QEye}

PostFusion-QEye, outlined in \Cref{fig:postfusion}, processes text and eye movements separately and combines their representations through two cross-attention mechanisms. The primary objective of these mechanisms is to transform both text and eye movement data into a unified space, which we refer to as the \textit{reading space} while taking into account the reading comprehension prediction task. 

The input paragraph is passed through a language model to obtain contextualized embeddings $Z_P$. The eye movement input features are processed through two 1D convolution layers, resulting in the eye movement representation $Z_{E_P}$. Cross-attention is then applied between the paragraph embedding $Z_P$ and $Z_{E_P}$, with eye movements as the query and text embeddings as the key and the value. This step modifies the paragraph words based on the eye movements. The output is provided along with $Z_{E_P}$ to a fully connected layer, yielding $Z_{E_P+P}$, a projection of the two into a shared space. Another cross-attention layer is applied between $Z_{E_P+P}$ as key and value and the question embedding $Z_Q$ as query, weighting the shared representation by the relevance to the question. The output of this step is passed to a multilayer perceptron classifier to predict the response.

\subsection{Multiple-Choice Variants}

For the specific-answer prediction task, we add to the model input the answer choices: $[a^{q^p}_1, a^{q^p}_2, a^{q^p}_3, a^{q^p}_4]$. The answer choices are provided to the model in a randomized order, as presented to the participants.

\subsection{Baseline Models}
\label{sec:baselines}

We compare the proposed models to a number of eye movement models from prior work. We focus on models that were either designed for reading comprehension prediction or can be adjusted to the binary task with minimal modifications. As none of the prior models allow encoding of answers, we cannot apply them to the multiple-choice task.

\paragraph{Logistic Regression} \cite{meziere2023using} Based on \citet{meziere2023using} who used linear regression for reading comprehension prediction. We use the same feature set which includes reading speed, and global averages of standard eye movement measures. 

\paragraph{CNN} \cite{Ahn2020SBSAT} Similarly to \citet{meziere2023using}, this model is based only on eye movement information, without the underlying text. It uses the fixation sequence, represented by x and y coordinates on the screen, fixation durations, and pupil size, which are passed through a Convolutional Neural Network (CNN) to predict a binary comprehension outcome.

\paragraph{BEyeLSTM} \cite{ReichETRA2022} A model for predicting reading comprehension from eye movements which represents both the fixation sequence and text features, combining LSTMs with affine transformations. BEyeLSTM outperforms the CNN model of \citet{Ahn2020SBSAT}, on the high versus low comprehension task with SB-SAT.

\paragraph{Eyettention} \cite{Deng_Eyettention2023} This model was originally developed for scanpath prediction. Eyettention is a word sequence encoder and a fixation sequence encoder that uses a pre-trained BERT \cite{devlin_bert_2019} and an LSTM \cite{hochreiter_long_1997}, with a cross-attention mechanism for the alignment of the input sequences. We adjust this model for prediction of reading comprehension by using global cross-attention instead of windowed attention, and represent the scanpath using the last hidden representation. Further details on this model are provided in \Cref{sec:app-baselines-mods}.

\subsection{No Eye Movements Baselines}
\label{sec:baselines2}

We further introduce two baselines with no eye movements. The first is a majority class baseline. The second is \textbf{Text-only RoBERTa}. This baseline is of special importance as it is able to take into account item difficulty as reflected in the item textual characteristics and the distribution of item responses in the training data. To our knowledge, no previous reading comprehension prediction method was benchmarked against this kind of baseline.

\begin{table*}[ht!]
\centering
\small
\resizebox{\textwidth}{!}{%
\begin{tabular}{@{}lllcS[table-format=2.2]ccS[table-format=2.2]ccc@{}} 
\toprule
\multicolumn{1}{l}{Binary Reading Comprehension} &  & & \multicolumn{4}{c}{Ordinary Reading (Gathering)} & \multicolumn{4}{c}{Information Seeking (Hunting)} \\ \cmidrule(lr){1-3} \cmidrule(lr){4-7} \cmidrule(lr){8-11}
Model & {\makecell{Gaze \\ Representation}} & {\makecell{Text \\ Representation}} & {\makecell{New \\ Item}} & {\makecell{New \\ Participant}} & {\makecell{New Item\\ \& Participant}} & {\makecell{All}} & {\makecell{New \\ Item}} & {\makecell{New \\ Participant}} & {\makecell{New Item\\ \& Participant}} & {\makecell{All}} \\
\midrule
Majority & None & None & 50.0 & 50.0 & 50.0 & 50.0 & 50.0 & 50.0 & 50.0 & 50.0 \\
Text-only RoBERTa & None & Emb &54.8 & 63.1 & 55.2 & 58.7 & 51.8 & 63.1 & 50.5 & 57.1 \\
\addlinespace[1ex]
Log. Reg. \cite{meziere2023using} & Global & None & 
53.3 & 50.8 & 53.8 & 52.2 & 53.2 & 52.2 & 52.3 & 52.7 \\
CNN \cite{Ahn2020SBSAT} & Fixations & None & 51.0 & 51.0 & 51.9 & 51.1 & 51.4 & 51.3 & 49.2 & 51.2 \\
BEyeLSTM \cite{ReichETRA2022}& Fixations & Ling. Feat. & 50.6 & 55.7 & 51.1 & 53.0 & 50.5 & 55.1 & \textbf{55.1} & 53.0 \\
Eyettention \cite{Deng_Eyettention2023} & Fixations & Emb + Word Len. & 54.8 & 60.4 & \textbf{57.1} & 57.6 & 50.5 & 56.4 & 52.3 & 53.4 \\
\addlinespace[1ex]
RoBERTa-QEye & Words & Emb + Ling. Feat. &\textbf{55.5} & 63.5 & 52.1 & 59.1 & 50.5 & \textbf{63.8} & 51.0 & 56.8 \\
RoBERTa-QEye & Fixations & Emb + Ling. Feat. & 53.3 & 61.3 & \textbf{57.1} & 57.3 & 50.3 & 60.3 & 50.8 & 55.1 \\
MAG-QEye & Words & Emb + Ling. Feat. & 54.8 & \textbf{64.1}*& 53.8 & \textbf{59.2} & 52.5 & 62.3 & 51.3 & 57.1 \\
PostFusion-QEye & Fixations & Emb + Ling. Feat. & 54.8 & 63.5 & 55.0 & 58.9 & \textbf{53.8}* & 62.7 & 53.8 & \textbf{58.0} \\
\bottomrule
\end{tabular}%
}
\caption{Results on balanced accuracy for the main binary reading comprehension prediction task (correct vs incorrect comprehension).  
`All' denotes results for the aggregation of all the trials across the three test regimes. `Emb' stands for word embeddings, `Ling. Feat.' for linguistic word properties. Statistically significant improvements over the Text-only RoBERTa baseline, using a paired bootstrap test, chosen based on considerations described in \citep{dror2018hitchhiker}, are marked with `*' at $p<0.05$.
}
\label{tab:results-combined}
\end{table*}

\begin{table*}[ht]
\centering
\resizebox{\textwidth}{!}{%
\begin{tabular}{@{}lllS[table-format=2.1,parse-numbers=false]S[table-format=2.1,parse-numbers=false]S[table-format=2.1,parse-numbers=false]S[table-format=2.1,parse-numbers=false]S[table-format=2.1,parse-numbers=false]S[table-format=2.1,parse-numbers=false]S[table-format=2.1,parse-numbers=false]S[table-format=2.1,parse-numbers=false]@{}}
\toprule
\multicolumn{3}{l}{Multiple-Choice Reading Comprehension}  & \multicolumn{4}{c}{Ordinary Reading (Gathering)} & \multicolumn{4}{c}{Information Seeking (Hunting)} \\ \cmidrule(lr){1-3} \cmidrule(lr){4-7} \cmidrule(lr){8-11}
Model & {\makecell{Gaze \\ Representation}} & {\makecell{Text \\ Representation}}  & {\makecell{New \\ Item}} & {\makecell{New \\ Participant}} & {\makecell{New Item \\\&  Participant}} & {\makecell{ \ \ \ \ \ All \ \ \ \ \ }} & {\makecell{New \\ Item}} & {\makecell{New \\ Participant}} & {\makecell{New Item \\ \&  Participant}} & {\makecell{ \ \ \ \ \ All \ \ \ \ \ }} \\ \midrule
Majority & None & None &  25.0& 25.0&  25.0& 25.0& 25.0&  25.0&  25.0&  25.0\\
Text-only RoBERTa & None & Emb &  25.3 & \textbf{33.0} & 25.2 & 29.0 & 25.0 & \textbf{31.7} & 24.8 & 28.2\\ \addlinespace[1ex]
RoBERTa-QEye & Words & Emb + Ling. Feat. & 28.2\textsuperscript{*} & 31.5 & 32.1\textsuperscript{**} & 29.9 & 28.9\textsuperscript{***}&30.1&27.1& 29.3 \\
RoBERTa-QEye & Fixations & Emb + Ling. Feat. & 29.2 \textsuperscript{*}& 32.9 & 28.1 & \textbf{30.9} & \textbf{30.3}\textsuperscript{***} & 31.0 &  \textbf{29.5} & \textbf{30.5}\textsuperscript{***} \\
MAG-QEye & Words & Emb + Ling. Feat. & 27.9\textsuperscript{***} & 32.5 & \text{30.4}\textsuperscript{***} & \text{30.2}\textsuperscript{**} & 26.8 & 30.0 & 29.0 & 28.4 \\ 
PostFusion-QEye & Fixations & Emb + Ling. Feat. &  \textbf{29.4}\textsuperscript{**} & 31.7 & \textbf{32.9}\textsuperscript{*}  & 30.6\textsuperscript{*} & 27.5\textsuperscript{*} & 27.9 & 26.7 & 27.6 \\ 
\bottomrule
\end{tabular}%
}
\caption{Results on balanced accuracy for the multiple-choice specific answer prediction task. Statistically significant improvements over the Text-only RoBERTa baseline, using a paired bootstrap test, are marked with `*' at $p<0.05$, `**' at $p<0.01$ and `***' at $p<0.001$. We note that in some cases, higher balanced accuracy scores correspond to lower p-values due to higher variability in the predictions of the minority classes.
}
\label{tab:results-extension}
\end{table*}

\section{Experimental Setup}

We evaluate the models in three evaluation regimes that test different aspects of model generalization. 
\begin{itemize}
    \item \textbf{New Participant}: No eyetracking data is available for the given participant, but eyetracking data from other participants is available for the given item (paragraph).
    \item \textbf{New Item}: No eyetracking data is available for the item, but prior eyetracking data is available for the participant on other items.
    \item \textbf{New Item \& Participant}: No prior eyetracking data is available for the participant nor for the item.
\end{itemize}
We further report aggregated results across all three regimes.

\begin{figure}[ht]
    \centering
\includegraphics[width=1\columnwidth]{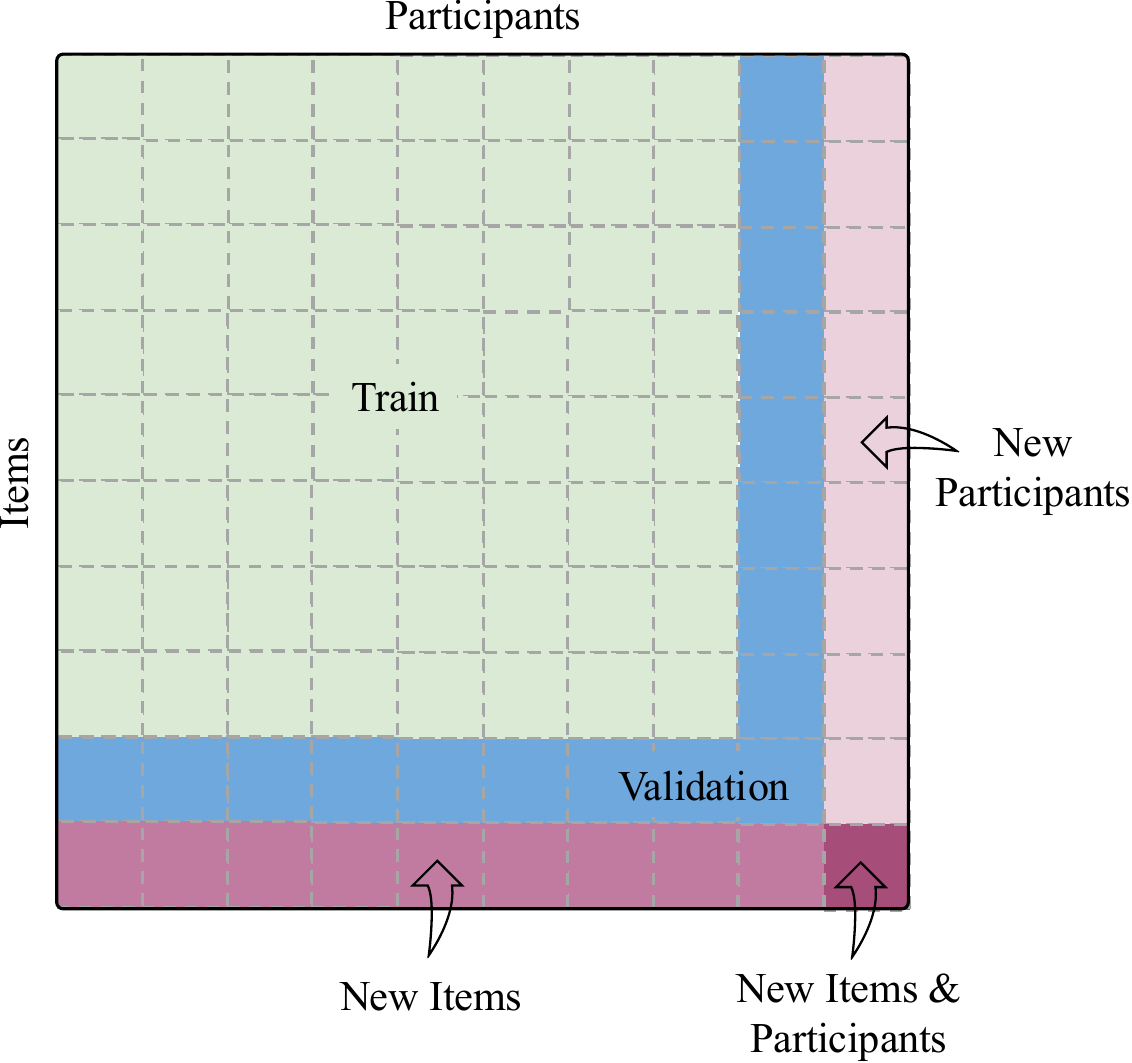}
    \caption{A schematic depiction of a 10-article 60-participant batch split, divided into a train set, a validation set, and the three test sets. A full data split for a reading regime (ordinary reading or information seeking) consists of the union of three batch splits. 
    }
    \label{fig:data-split}
\end{figure}

We perform model training, hyperparameter tuning, and evaluation separately for the ordinary reading and information seeking parts of the data, with $10$-fold cross-validation. \Cref{fig:data-split} presents schematically one of the 10 data splits for a 10-article 60-participant batch. A full data split for a reading regime (ordinary reading or information seeking) is the union of three such splits. In each split, approximately 64\% of the data is allocated for training, 17\% for validation, and 19\% for testing. The test data is further divided into 9\% in the New Participant, 9\% New Item, and 1\% New Item \& Participant regimes. In total across the 10 splits, approximately 90\% of the trials in the dataset appear in each of the New Participant and New Item evaluation regimes, and 10\% in the New Item \& Participant regime. 
Items are assigned to the train, validation and test portions of each split at the \emph{article level}, such that no article is split across different data portions, ensuring generalization to items whose content is unrelated to items seen in training. See \Cref{sec:app-cv-splits} for further information on the splits. 

Because the data is unbalanced across classes, we use balanced accuracy as the evaluation metric. As prior work has shown considerable differences in reading behavior between the ordinary reading and information seeking reading conditions \cite{hahn_modeling_2023,malmaud_bridging_2020,shubi2023cogsci}, we train and evaluate the models on each type of trials separately. We perform hyperparameter tuning for each split, and report balanced accuracy results on the aggregation of the predictions across the 10 test sets. We assume that at test time the evaluation regime of the trial is \emph{unknown}. Model hyperparameter tuning is therefore based on the entire validation set of the split.
As prior models from the literature were developed for different tasks and on different datasets, we run a hyperparameter search for each model over a search space that includes the original parameter settings. Hyperparameters are also optimized for the Text-only RoBERTa baseline. To address the unbalanced nature of the data, shown in \Cref{tab:starc}, we sample the same number of trials from each answer class during training. Additional details on feature normalization, model training, hyperparameter search, and number of model parameters are provided in \Cref{sec:app-hyperparams}.

\section{Results}
\subsection{Correct vs Incorrect Comprehension}

In \Cref{tab:results-combined}, we present trial-level reading comprehension prediction results for ordinary reading and information seeking. The best results are achieved by different models under the different evaluation regimes. MAG-QEye achieves the highest overall balanced accuracy in ordinary reading with a score of 59.2, while PostFusion-QEye performs best in information seeking, with a score of 58.0. In all the evaluation regimes, the best performing model outperforms the Text-only RoBERTa baseline. In all but the New Item \& Participant evaluation regime, the best performing model is one of our proposed models. Text-only RoBERTa turns out to be a key benchmark, whereby most models are below this baseline especially in the New Participant regime.

We note several key trends in the results. First, results in the New Participant regime tend to be higher than in the New Item regime, highlighting the importance and the challenge of generalization to new items. The strong performance of the RoBERTa text-only baseline in the New Participant regime suggests that much of the gains in this regime do not stem from eye movement information, but rather from item properties and statistics. This highlights the importance of benchmarking against such a baseline for assessing the contribution of eye movement information. 
It further underscores the importance of explicit representation of the text; the Logistic Regression, CNN and BEyeLSTM models, which do not include such a representation, perform poorly in the New Participant regime. Finally, for any given model, the ordinary reading regime tends to yield higher accuracies compared information seeking. We hypothesize that this difference could be related to higher variability in reading strategies in information seeking across participants \cite{shubi2023cogsci}. We leave a detailed investigation of this hypothesis to future work. 

\subsection{Multiple-Choice Task}

In \Cref{tab:results-extension} we use our models, MAG-QEye and PostFusion-QEye, and the two RoBERTa-QEye variants to predict  participants' specific answer response among the four provided answers. As mentioned above, prior models from the literature are not applicable for this task. We find that all the models outperform the Text-only RoBERTa baseline in the two regimes that involve new items, but not in the New Participant regime. The best performing model in the overall evaluations is RoBERTa-QEye-Fixations. The general trends regarding higher performance in the New Participant regime compared to the New Item regime, as well as the stronger within-model performance in ordinary reading compared to information seeking, extend to this evaluation.

\subsection{Additional Experiments}

We perform two additional sets of experiments of preliminary nature.
In \Cref{sec:app-eyelin} we provide ablation experiments on the effect of linguistic word properties on model performance. In \Cref{sec:app-backbone} we further examine different variants of the textual backbone of the models. Finally, we provide validation set results in \Cref{sec:app-val-results}.

\section{Summary and Discussion}

This paper presents a systematic evaluation of the ability to predict reading comprehension from eye movements in reading at the level of a single question over a single paragraph. We address this task using a range of existing and new models, applied to large scale data across several task variants and evaluation regimes. Our experiments indicate that the task at hand is highly challenging, and further highlight the importance of text-only baselines for assessing the added value of eye movements information. However, we do find that small improvements over a strong text-only baseline are achievable with the proposed and some of the past modeling approaches. 

Given the presented results, the extent to which specific aspects of reading comprehension can be reliably decoded from eye movements signal remains an open question. It is possible that eye movements simply do not contain sufficient information for decoding comprehension at high accuracy rates for the examined level of  granularity. Alternatively, it may be the case that current modeling techniques do not represent or process eye movements data effectively enough for this task. Another factor whose role in task difficulty needs to be investigated in more detail is the imbalanced nature of the data, where only a relatively small fraction of the responses are incorrect. 

Additional work on eye movement data analysis, new model architectures, feature representations and training regimes is needed  for making further progress on this task. Additionally, new datasets with other task variants and other populations such as children and L2 readers are required to study the problem in a more comprehensive manner. We envision that the models, tasks, evaluation protocols, and data presented here will serve as a stepping stone for such work, as well as a broader scientific investigation of the relations between eye movements and reading comprehension.

\section{Ethical Considerations}

The eyetracking data used in this work was collected by \citet{malmaud_bridging_2020} under an institutional IRB protocol. All the participants provided written consent prior to participating in the eyetracking study. The data is anonymized. 
Analyses of the relations between eye movements and reading comprehension, and predictive models of comprehension are among the primary use cases for which the data was collected. 

Automatic reading comprehension assessments from eye movements can potentially address shortcomings of standard assessment methodologies by reducing test development and test taking costs, and enhancing test availability. However, they also introduce potential risks for biased and inaccurate assessments that may put various populations and individuals at a disadvantage. These include non-native speakers, older participants, participants with cognitive impairments, disabilities, eye conditions and others. Much higher model performance than the current state-of-the-art and a thorough examination of potential biases due to factors unrelated to reading comprehension are needed before considering deploying such assessments. 

It has previously been shown that eye movements can  be used for user identification \cite[e.g.][]{bednarik2005eye,jager2020deep}. We do not perform user identification in this study. 
We further emphasize that future reading comprehension assessment systems are to be used only with explicit consent from potential users to have their eye movements collected and analyzed for this purpose.

\section{Limitations}
\label{sec:limitations}

Our work has a number of limitations which are related to the experimental design of OneStop. First, the textual data consists of articles with 4-7 paragraphs. Each question is over the content of a single paragraph. Longer and shorter texts, as well as questions that require integration of information from several paragraphs, are not covered. The experimental design does not allow participants to go back and forth between the question and passage, which is common in question answering tasks. Further, participant expectations for upcoming reading comprehension questions, as well as the setting of an in-lab experiment may result in reading patterns that deviate from reading in everyday settings \citep{huettig_myth_2022} and could impact the predictive performance of the model.

While our work examines the feasibility of automated assessment of reading comprehension from eye movements, the accuracy of the models presented is still very far from being relevant for deployment in real world scenarios. Our results are further limited to the equipment at hand. Our approach has only been tested using a state-of-the-art eyetracker (Eyelink 1000 Plus) at a sampling rate of 1000Hz. This allows  extracting gaze position and duration at a very high temporal resolution and character-level precision. While studies such as  \citet{ishimaru_towards_2017} and \citet{chen_characteristics_2023} have demonstrated predictive modeling capabilities using lower spatial and temporal resolution eye tracking systems, additional work is required to test the feasibility of reading comprehension prediction using such equipment. 

Although we use the largest eyetracking for reading comprehension dataset to date, OneStop was collected from adult L1 English speakers, with no cognitive impairments, and in the large majority of cases no eye conditions. We acknowledge that this pool of participants excludes multiple populations, including children, elderly, participants with cognitive and physical impairments and others. Future data collection and analysis work is required to test the generalization capabilities and potential biases of the models in other populations.

In this work we assume the availability of both suitable eyetracking data and a pretrained language model for the language at hand. Although language models for lower-resource languages \cite[e.g.][]{chriqui_hebert_2022,vamvas2023swissbert} and multilingual models \cite[e.g.][]{lai-etal-2023-chatgpt} have been made available, many languages still lack such models. Similarly, to the best of our knowledge, no eyetracking data with a substantial reading comprehension component is currently available for languages other than English. This limits the generality of the results. 
More eyetracking data collection and language model development work is required to include additional languages.

\section*{Acknowledgments}
This work was supported by ISF grant 1499/22.

\bibliography{custom,references_zotero} 

\clearpage

\appendix

\section{Features}
\label{sec:app-eye-features}

\begin{table}[H]
\onecolumn
\centering
\resizebox{\textwidth}{!}{%
\begin{tabular}{@{}ll@{}}
\toprule
\multicolumn{1}{c}{\multirow{2}{*}{\textbf{Feature Name}}} & \multicolumn{1}{c}{\multirow{2}{*}{\textbf{Description}}} \\
\multicolumn{1}{c}{} & \multicolumn{1}{c}{} \\ \midrule
\textbf{Word-Level Eye Movement Features} & \\
\midrule
IA\_DWELL\_TIME & The sum of the duration across all fixations that fell in the current interest area \\
IA\_DWELL\_TIME\_\% & Percentage of trial time spent on the current interest area (IA\_DWELL\_TIME / TRIAL\_DWELL\_TIME). \\
IA\_FIXATION\_\% & Percentage of all fixations in a trial falling in the current interest area. \\
IA\_FIXATION\_COUNT & Total number of fixations falling in the interest area. \\
IA\_REGRESSION\_IN\_COUNT & Number of times interest area was entered from a higher IA\_ID (from the right in English). \\
IA\_REGRESSION\_OUT\_FULL\_COUNT & Number of times interest area was exited to a lower IA\_ID (to the left in English). \\
IA\_RUN\_COUNT & Number of times the Interest Area was entered and left (runs). \\
IA\_FIRST\_FIX\_PROGRESSIVE & Checks whether the first fixation in the interest area is a first-pass fixation. \\
IA\_FIRST\_FIXATION\_DURATION & Duration of the first fixation event that was within the current interest area \\
IA\_FIRST\_FIXATION\_VISITED\_IA\_COUNT & This reports the number of different interest areas visited so far before the first fixation is made to the current interest area. \\
IA\_FIRST\_RUN\_DWELL\_TIME & Dwell time of the first run (i.e., the sum of the duration of all fixations in the first run of fixations within the current interest area). \\
IA\_FIRST\_RUN\_FIXATION\_COUNT & Number of all fixations in a trial falling in the first run of the current interest area. \\
IA\_SKIP & An interest area is considered skipped (i.e., IA\_SKIP = 1) if no fixation occurred in first-pass reading. \\
IA\_TOP & Y coordinate of the top of the interest area. \\
IA\_LEFT & X coordinate of the left-most part of the interest area. \\
normalized\_Word\_ID & Position in the paragraph of the word interest area, normalized from zero to one. \\
IA\_REGRESSION\_PATH\_DURATION & The summed fixation duration from when the current interest area is first fixated until  the eyes enter an interest area with a higher IA\_ID. \\
IA\_REGRESSION\_OUT\_COUNT & Number of times interest area was exited to a lower IA\_ID (to the left in English) before a higher IA\_ID was fixated in the trial. \\
IA\_SELECTIVE\_REGRESSION\_PATH\_DURATION & Duration of fixations and refixations of the current  interest area before the eyes enter an interest area with a higher ID. \\
IA\_LAST\_FIXATION\_DURATION & Duration of the last fixation event that was within the current interest area. \\
IA\_LAST\_RUN\_DWELL\_TIME & Dwell time of the last run (i.e., the sum of the duration of all fixations in the last run of fixations within the current interest area). \\
PARAGRAPH\_RT & Reading time of the entire paragraph. \\
total\_skip & Binary indicator whether the word was fixated on. \\
\midrule
\textbf{Fixation-level Eye Movement Features} & \\
\midrule
CURRENT\_FIX\_INDEX & The position of the current fixation in the trial. \\
CURRENT\_FIX\_DURATION & Duration of the current fixation. \\
CURRENT\_FIX\_PUPIL & Average pupil size during the current fixation. \\
CURRENT\_FIX\_X & X coordinate of the current fixation. \\
CURRENT\_FIX\_Y & Y coordinate of the current fixation. \\
NEXT\_FIX\_ANGLE, PREVIOUS\_FIX\_ANGLE & Angle between the horizontal plane and the line connecting the current fixation and the next/previous fixation. \\
NEXT\_FIX\_DISTANCE, PREVIOUS\_FIX\_DISTANCE & Distance between the current fixation and the next/previous fixation in degrees of visual angle. \\
NEXT\_SAC\_AMPLITUDE & Amplitude of the following saccade in degrees of visual angle. \\
NEXT\_SAC\_ANGLE & Angle between the horizontal plane and the direction of the next saccade. \\
NEXT\_SAC\_AVG\_VELOCITY & Average velocity of the next saccade. \\
NEXT\_SAC\_DURATION & Duration of the next saccade in milliseconds. \\
NEXT\_SAC\_PEAK\_VELOCITY & Peak values of gaze velocity (in visual degrees per second) of the next saccade. \\
\bottomrule
\end{tabular}%
}
\caption{Word-level and fixation-level eye movement features, defined and extracted by SR Data Viewer.}
\label{tab:combined-eye-movement-features}
\end{table}

\begin{table}[ht]
\centering
\resizebox{\textwidth}{!}{%
\begin{tabular}{@{}ll@{}}
\toprule
\multicolumn{1}{c}{\textbf{Feature Name}} & \multicolumn{1}{c}{\textbf{Description}} \\
\midrule
Surprisal &  \makecell[l]{\cite{hale2001probabilistic,levy2008expectation}, formulated as $-\log_2(p(word|context))$ for each \textit{word} given the preceding textual content of the \\ paragraph as \textit{context}, probabilities extracted from the GPT-2-small language model \cite{radford2019language,wolf-etal-2020-transformers}.} \\
Wordfreq\_Frequency & Frequency of the word based on the Wordfreq package \cite{robyn_speer_2022_7199437}, formulated as $-\log_2(p(word))$. \\
Length & Length of the word in characters. \\
start\_of\_line & Binary indicator of whether the word appeared at the beginning of a line. \\
end\_of\_line & Binary indicator of whether the word appeared at the end of a line. \\
Is\_Content\_Word & \makecell[l]{Binary indicator of whether the word is a content word. \\ A content word is defined as a word that has a part-of-speech tag of either PROPN, NOUN, VERB, ADV, or ADJ.} \\
n\_Lefts & The number of leftward immediate children of the word in the syntactic dependency parse.\\
n\_Rights & The number of rightward immediate children of the word in the syntactic dependency parse.

\\
Distance2Head & The number of words to the syntactic head of the word.\\
\bottomrule
\end{tabular}%
}
\caption{Linguistic word properties and their descriptions. POS tags and parse trees were obtained using SpaCy \cite{Honnibal2020}.}
\label{tab:additional-features}
\end{table}

\clearpage

\newpage

\section{Adaptations of Prior Models}
\label{sec:app-baselines-mods}

\subsection{MAG}
\label{sec:app-baselines-mods-mag}

We replace the vision and acoustic input with word-level eye movement features. To align them with the tokenized text, we duplicate the word-level features for each subword token.
Additionally, for a fair comparison with other models, we replace BERT with $\text{RoBERTa}_{\text{LARGE}}$ as the textual backbone model. 

Formally, each token embedding $Z_i$ is \textit{displaced} by $H_i$. 
\begin{equation}
    \bar{Z_i} = Z_i + \alpha H_i
\end{equation}

$H_i$ is a scaled and transformed version of the eye movements $E_i$,
\begin{equation}
    H_i = g_i \cdot (W_e E_i)+b_H
\end{equation}

where the scaling is defined by,
\begin{equation}
    g_i = ReLU(W_g[Z_i; A_i] + b_g)
\end{equation}

The amount of displacement is defined by
\begin{equation}
    \alpha=min(\frac{|| Z_i||_2}{|| H_i||_2}\beta,1)
\end{equation}

where $\beta$ is a hyper-parameter, and $W_e,W_g, b_H, b_g$ are learned.

Finally, the contextualized \texttt{CLS} token is used for classification.

\subsection{Eyettention}
We adjust the prediction objective of the model from next fixation to trial-level classification. To this end, we use global cross attention between the word sequence and the scanpath sequence instead of fixed window cross attention, as suggested in \citet{Deng_Eyettention2023}. We then represent the whole scanpath using the last hidden representation of the scanpath LSTM. We further replace BERT, with $\text{RoBERTa}_{\text{LARGE}}$ for consistency with the other models.

\subsection{BEyeLSTM}
First, we employ SpaCy tokenization based on paragraph-level input rather than word-level input, resulting in a more precise tokenization. Second, the textual materials used here include a more fine-grained set of part-of-speech tags and named entities, which results in a larger final feature set. Lastly, we omit the "words in fixed context on unigrams" feature, as it presupposes that all the participants read the same texts, which is not the case in OneStop.

\subsection{CNN} 
\citet{Ahn2020SBSAT} resort to artificially subdividing SB-SAT texts into smaller segments in order generate a sufficient number of training examples to make the dataset usable for their task of predicting low versus high comprehension over multiple items. This heuristic is problematic in general, and not applicable to the single item task addressed here. In the current work we use the entire fixation sequence as the input to the model. 

\newpage

\section{Cross Validation Splits}
\label{sec:app-cv-splits}

Each split guarantees an equal number of participants from each OneStopQA batch in each portion of the split, and is approximately stratified by answer type. 
Recall that each participant is presented with a specific combination of a paragraph and one of its three associated questions. Due to the stratification by answer type, it is not guaranteed that the appearances of any given paragraph will be balanced across the three possible questions in any of the split portions.
Note that across the 10 test sets, not all participant -- item combinations are covered in the test sets, as this would require 100 data splits.

\newpage

\section{Feature Standardization and Hyperparameter Tuning}
\label{sec:app-hyperparams}

We apply standardization for each feature in $E_P$, where the statistics are computed on the train set and applied to the validation and test sets, separately for each split. Feature normalization is performed using Scikit-learn \cite{pedregosa_scikit-learn_2011}.

For all the neural models, we use the AdamW optimizer \cite{loshchilov_decoupled_2018} with a batch size of $16$, a linear warmup ratio of $0.1$, and a weight decay of 0.1, following best practice recommendations from \citet{liu_roberta_2019} and \citet{mosbach2021on}. The search space for learning rates is $\{0.00001, 0.00003, 0.0001\}$ and for dropout $\{0.1, 0.3, 0.5\}$. 

\begin{itemize}
    \item For \textbf{Logistic Regression}, we search over regularization parameter C values of $\{0.1, 5, 10, 50, 100\}$, with and without an L2 penalty.
    \item For the \textbf{CNN} we include a learning rate of 0.001 as in \citet{Ahn2020SBSAT}.
    \item Following \citep{ReichETRA2022}, for \textbf{BEyeLSTM} the search space for learning rates is $\{0.001, 0.003, 0.01\}$, embedding dimensions of $\{4,8\}$ and hidden dimension of $\{64,128\}$.
    \item For \textbf{Eyettention} we also include a learning rate of 0.001 and dropout of 0.2, as in \citet{Deng_Eyettention2023}. 
    \item For \textbf{MAG-QEye}, the search space for the injection layer index is $\{0, 11, 23\}$. We set the MAG-internal dropout to 0.5, and the scaler parameter to 1e-3, as suggested by \cite{rahman_integrating_2020}. 
    \item  In \textbf{PostFusion-QEye}, the 1D convolution layers have a kernel size of three, stride 1, and padding 1.
\end{itemize}

All neural networks are trained using the Pytorch Lighting library \cite{Falcon_PyTorch_Lightning_2019,paszke2019pytorch} and evaluated using torch-metrics \cite{TorchMetrics_-_Measuring_2022} on a NVIDIA A100-40GB and A40-48GB GPUs.  We adapt Huggingface's RoBERTa implementation \cite{wolf-etal-2020-transformers}.
The baselines described in \Cref{sec:baselines} are reimplemented in this framework as well.  A single training epoch took approximately 5 minutes. We train for a maximum of ten epochs, stopping after three epochs without improvement on the validation set.

The number of model parameters is 355M for the $\text{RoBERTa}_{\text{LARGE}}$ backbone, and an additional 1.1M for MAG-QEye and RoBERTa-QEye, and 9M for PostFusion-QEye.

\newpage 

\section{The Role of Linguistic Word Property Features}
\label{sec:app-eyelin}

Our proposed models tend to outperform the Text-only RoBERTa baseline, especially in the two evaluation regimes that involve new items. Note however, that in addition to eye movements, these models also include linguistic word properties, which may provide information on the textual item that is not fully encoded in word embeddings. Some of them (e.g. word length, frequency and surprisal) are also known to be predictive of reading times. 

What is the effect of these features on model performance? To examine this question, we carry out two ablation experiments. In the first experiment, we ablate the linguistic word property features. In the second experiment we ablate the eye movement features. The latter ablation is not possible with fixation based models, because even with the eye movement features removed, these models still have information about the gaze trajectory through the order and word identity of the fixations. We therefore perform these experiments only with the word based models RoBERTa-QEye-Words and MAG-QEye.

\Cref{tab:results-input-ablation-binary} in \Cref{sec:app-eyelin} presents the ablation results for the binary task. In the first experiment, removal of linguistic word properties does not substantially affect model performance. This outcome does not match our expectation regarding the potential benefits of allowing models to learn eye movement -- linguistic word property interactions. In the second experiment, overall, we again do not observe performance degradation when ablating the eye movement features. While this experiment is not sufficient for drawing general conclusions regarding the value of eye movement information for our task, it suggests that in our two instances of word-based models, eye movements do not seem to provide substantial performance gains above and beyond features that can be readily extracted from the text. We leave a more extensive investigation regarding the impact of linguistic features on model performance to future work.

\begin{table}[ht]
\centering
\resizebox{\columnwidth}{!}{%
\begin{tabular}{@{}l
S[table-format=2.1,parse-numbers=false]
S[table-format=2.1,parse-numbers=false]
S[table-format=2.1,parse-numbers=false]
S[table-format=2.1,parse-numbers=false]
S[table-format=2.1,parse-numbers=false]
S[table-format=2.1,parse-numbers=false]
S[table-format=2.1,parse-numbers=false]
S[table-format=2.1,parse-numbers=false]@{}}
\toprule
\multicolumn{1}{l}{Binary Reading Comprehension} & \multicolumn{4}{c}{Gathering Trials} & \multicolumn{4}{c}{Hunting Trials} \\  \cmidrule(lr){1-1} \cmidrule(lr){2-5} \cmidrule(lr){6-9}
Model & {\makecell{New \\ Item}} & {\makecell{New \\ Participant}} & {\makecell{New Item \\\&  Participant}} & {\makecell{All}} & {\makecell{New \\ Item}} & {\makecell{New \\ Participant}} & {\makecell{New Item \\ \&  Participant}} & {\makecell{All}} \\ \midrule
Text-only RoBERTa & 54.8 & 63.1 & 55.2 & 58.7 & 51.8 & 63.1 & 50.5 & 57.1 \\
 \midrule
MAG-QEye  & 54.8 & \textbf{64.1*} & 53.8 & 59.2 & \textbf{52.5} & 62.3 & 51.3 & 57.1 \\
MAG-QEye w/o Ling. Feat & 55.9 & 63.8 & 55.5 & 59.6 & 52.3 & 63.3 & \textbf{54.8} & \textbf{57.7} \\
MAG-QEye w/o Eyes & 54.2 & 63.7 & 56.7 & 58.8 & 51.9 & 63.3 & 53.8 & 57.4 \\
 \midrule
RoBERTa-QEye-Words & 55.5 & 63.5 & 52.1 & 59.1 & 50.5 & \textbf{63.8} & 51.0 & 56.8  \\
RoBERTa-QEye-Words w/o Ling. Feat & 55.4 & 63.3 & 56.3 & 59.2 & 51.1 & 62.7 & 50.7 & 56.6 \\
RoBERTa-QEye-Words w/o Eyes & \textbf{56.7}\textsuperscript{*} & 63.7 & \textbf{57.5}& \textbf{60.0}\textsuperscript{**} & 49.3 & 63.2 & 51.2 & 56.0 \\
\bottomrule

\end{tabular}%
}
\caption{The effect of ablating word-level eye movement features (\Cref{tab:combined-eye-movement-features}) and linguistic word properties (\Cref{tab:additional-features}) on balanced accuracy for binary classification of the word based models MAG-QEye and RoBERTa-QEye-Words. 
Statistically significant improvements over Text-only RoBERTa, using a paired bootstrap test, are marked with `*' at $p<0.05$, `**' at $p<0.01$ and `***' at $p<0.001$.}
\label{tab:results-input-ablation-binary}
\end{table}

\newpage

\section{Textual Backbone Variants}
\label{sec:app-backbone}

Our models use RoBERTa as a textual backbone model, and the parameters of this backbone are subjected to change during model training. Other choices for this model component are possible. For example, one can pre-train the model on multiple choice question answering, freeze the textual backbone parameters during model training, or choose a different textual backbone model altogether. Preliminary experiments with MAG-QEye in \Cref{sec:app-backbone} \Cref{tab:results-backbone-ablation} do not show a consistent effect of these choices on model performance in the main prediction task. We leave a comprehensive investigation of textual backbone model choice and training to future work.

\begin{table}[h]
\centering
\resizebox{\columnwidth}{!}{%
\begin{tabular}{@{}lS[table-format=2.1,parse-numbers=false]
S[table-format=2.1,parse-numbers=false]
S[table-format=2.1,parse-numbers=false]
S[table-format=2.1,parse-numbers=false]
S[table-format=2.1,parse-numbers=false]
S[table-format=2.1,parse-numbers=false]
S[table-format=2.1,parse-numbers=false]
S[table-format=2.1,parse-numbers=false]@{}}
\toprule
 \multicolumn{1}{l}{Binary Reading Comprehension} & \multicolumn{4}{c}{Gathering Trials} & \multicolumn{4}{c}{Hunting Trials} \\ \cmidrule(lr){1-1} \cmidrule(lr){2-5} \cmidrule(lr){6-9}
 MAG-QEye Backbone & {\makecell{New \\ Item}} & {\makecell{New \\ Participant}} & {\makecell{New Item \\\&  Participant}} & {\makecell{All}} & {\makecell{New \\ Item}} & {\makecell{New \\ Participant}} & {\makecell{New Item \\ \&  Participant}} & {\makecell{All}} \\ \midrule
 RoBERTa Large & \textbf{54.8} & 64.1 & 53.8 & 59.2 & \textbf{52.5} & 62.3 & 51.3 & \textbf{57.1} \\
RoBERTa Large Frozen & 54.3 & 61.4 & 51.4 & 57.5 & 51.9 & 60.0 & \textbf{53.3} & 55.8 \\
 RoBERTa Large Trained for QA on RACE & \textbf{54.8} & \textbf{64.6} & 52.7 & \textbf{59.3} & 48.3 & 62.7 & 44.9 & 54.9 \\
RoBERTa Base & 52.8 & 64.0 & \textbf{56.9} & 58.3 & 50.8 & \textbf{63.5}\textsuperscript{*} & 51.6 & 56.9 \\ 
\bottomrule
\end{tabular}%
}
\caption{
Balanced accuracy performance comparison of different backbone architectures and training strategies for MAG-QEye. Statistically significant improvements compared to an unfrozen RoBERTa Large backbone are marked with `*' at $p<0.05$, `**' at $p<0.01$ and `***' at $p<0.001$ using a paired bootstrap test.}
\label{tab:results-backbone-ablation}
\end{table}

\newpage

\twocolumn

\section{Validation Set Results}
\label{sec:app-val-results}

\begin{table*}[ht!]
\onecolumn

\centering
\small
\resizebox{\textwidth}{!}{%
\begin{tabular}{@{}lllcccccccS[table-format=2.1,parse-numbers=false]@{}}
\toprule
\multicolumn{1}{l}{Binary Reading Comprehension} &  & & \multicolumn{4}{c}{Ordinary Reading (Gathering)} & \multicolumn{4}{c}{Information Seeking (Hunting)} \\ \cmidrule(lr){1-3} \cmidrule(lr){4-7} \cmidrule(lr){8-11}
Model & {\makecell{Gaze \\ Representation}} & {\makecell{Text \\ Representation}} & {\makecell{New \\ Item}} & {\makecell{New \\ Participant}} & {\makecell{New Item\\ \& Participant}} & {\makecell{All}} & {\makecell{New \\ Item}} & {\makecell{New \\ Participant}} & {\makecell{New Item\\ \& Participant}} & {\makecell{All}} \\
\midrule
Majority & None & None & 50.0 & 50.0 & 50.0 & 50.0 & 50.0 & 50.0 & 50.0 & 50.0 \\
Text-only RoBERTa & None & Emb & 59.8 & \textbf{65.8} & 57.9 & 62.5 & 57.1 & 65.1 & 56.8 & 60.8 \\
\addlinespace[1ex]
Log. Reg. \cite{meziere2023using} & Global & None & 53.4 & 51.1 & 53.9 & 52.3 & 51.8 & 53.0 & 51.9 & 52.4 \\
CNN \cite{Ahn2020SBSAT} & Fixations & None & 53.3 & 53.7 & 53.4 & 53.5 & 55.1& 54.5 & 55.0 & 54.8 \\
BEyeLSTM \cite{ReichETRA2022}& Fixations & Ling. Feat. & 55.0 & 58.5 & 55.7& 56.7 & 57.3 & 58.6 & 58.3 & 58.0 \\
Eyettention \cite{Deng_Eyettention2023} & Fixations & Emb + Word Len. & 58.5 & 62.4 & 57.9 & 60.3 & 57.0 & 59.5 & 56.9 & 58.2 \\
\addlinespace[1ex]
RoBERTa-QEye & Words & Emb + Ling. Feat. & 57.0 & 65.5 & \textbf{60.5} & 61.2& 55.3 & 64.7 & 52.2 & 59.6 \\
RoBERTa-QEye & Fixations & Emb + Ling. Feat. & 57.0 & 63.5 & 60.4 & 60.3 & 54.6 & 62.4 & 56.5 & 58.4 \\
MAG-QEye & Words & Emb + Ling. Feat. & \textbf{60.4} & \textbf{65.8} & 58.9 & \textbf{62.9} & 57.3 & \textbf{66.0} & \textbf{59.5} & 61.6 \\
PostFusion-QEye & Fixations & Emb + Ling. Feat. & 60.1 & 65.2 & 60.4 & 62.5 & \textbf{58.3} & 65.8 & 59.3 & \textbf{61.9}\textsuperscript{*} \\
\bottomrule
\end{tabular}%
}
\caption{Balanced accuracy for the binary reading comprehension prediction task (correct vs incorrect comprehension).
}
\label{tab:results-combined-val}
\end{table*}

\begin{table*}[ht]
\centering
\resizebox{\textwidth}{!}{%
\begin{tabular}{@{}lllS[table-format=2.1,parse-numbers=false]S[table-format=2.1,parse-numbers=false]S[table-format=2.1,parse-numbers=false]S[table-format=2.1,parse-numbers=false]S[table-format=2.1,parse-numbers=false]S[table-format=2.1,parse-numbers=false]S[table-format=2.1,parse-numbers=false]S[table-format=2.1,parse-numbers=false]@{}}
\toprule
\multicolumn{3}{l}{Multiple-Choice Reading Comprehension}  & \multicolumn{4}{c}{Ordinary Reading (Gathering)} & \multicolumn{4}{c}{Information Seeking (Hunting)} \\ \cmidrule(lr){1-3} \cmidrule(lr){4-7} \cmidrule(lr){8-11}
Model & {\makecell{Gaze \\ Representation}} & {\makecell{Text \\ Representation}}  & {\makecell{New \\ Item}} & {\makecell{New \\ Participant}} & {\makecell{New Item \\\&  Participant}} & {\makecell{ \ \ \ \ \ \ All \ \ \ \ \ \ }} & {\makecell{New \\ Item}} & {\makecell{New \\ Participant}} & {\makecell{New Item \\ \&  Participant}} & {\makecell{ \ \ \ \ \ \ All \ \ \ \ \ \ }} \\ \midrule
Majority & None & None & 25.0 & 25.0 & 25.0 & 25.0 & 25.0 & 25.0 & 25.0 & 25.0\\
Text-only RoBERTa & None & Emb & 25.7 & 35.7 & 25.6 & 30.4 & 25.0 & \textbf{34.4} & 25.5 & 29.5\\ \addlinespace[1ex]
RoBERTa-QEye & Words & Emb + Ling. Feat. & 34.0\textsuperscript{***} & 34.4 & 37.4\textsuperscript{**} & 34.3\textsuperscript{***} & 33.3 \textsuperscript{***}& 34.3 & 32.9 & 33.7\textsuperscript{*} \\ 
RoBERTa-QEye & Fixations & Emb + Ling. Feat. & 33.6\textsuperscript{***} & 34.7 &  \textbf{37.9}\textsuperscript{***} & 34.3\textsuperscript{***} & 34.0\textsuperscript{***} &  \textbf{34.4} &  \textbf{37.4} &  \textbf{34.3}\textsuperscript{***} \\ 
MAG-QEye & Words & Emb + Ling. Feat. & \textbf{33.8}\textsuperscript{***} & \textbf{36.1} & 34.3\textsuperscript{**} & \textbf{34.9}\textsuperscript{**} & \textbf{34.8}\textsuperscript{***} & 33.6 & 32.9 & 34.1\textsuperscript{***}\\ 
PostFusion-QEye & Fixations & Emb + Ling. Feat. & 33.2\textsuperscript{***} & 35.1 & 33.5\textsuperscript{*}  & 34.1\textsuperscript{**} & 34.0\textsuperscript{**} & 31.8 & 35.4 & 33.0 \\ 
\bottomrule
\end{tabular}%
}
\caption{Balanced accuracy for the multiple-choice specific answer prediction task.}
\label{tab:results-extension-val}
\end{table*}
\twocolumn

\end{document}